# FAST SAMPLING GENERATIVE MODEL FOR PLANE WAVE ULTRASOUND IMAGE RECONSTRUCTION

*Hengrong Lan[†], Zhiqiang Li[†], Qiong He, and Jianwen Luo[*]*

Department of Biomedical Engineering, School of Medicine, Tsinghua University, Beijing, China
[*]E-mail: luo_jianwen@tsinghua.edu.cn

## ABSTRACT

Image reconstruction from radio-frequency data is pivotal in ultrafast plane wave ultrasound imaging. Unlike the conventional delay-and-sum (DAS) technique, which relies on somewhat imprecise assumptions, deep learning-based methods perform image reconstruction by training on paired data, leading to a notable enhancement in image quality. Nevertheless, these strategies often exhibit limited generalization capabilities. Recently, denoising diffusion models have become the preferred paradigm for image reconstruction tasks. However, their reliance on an iterative sampling procedure results in prolonged generation time. In this paper, we propose a novel sampling framework that concurrently enforces data consistency of ultrasound signals and data-driven priors. By leveraging the advanced diffusion model, the generation of high-quality images is substantially expedited. Experimental evaluations on an *in-vivo* dataset indicate that our approach with a single plane wave surpasses DAS with spatial coherent compounding of 75 plane waves.

## 1. INTRODUCTION

Ultrasound (US) imaging is prevalently employed in medical diagnostics, primarily due to its portability, real-time capability, nonionizing radiation, and cost-effectiveness. In recent years, plane wave US (PWUS) imaging has garnered significant interest, given its capability to achieve frame rates up to 10,000 Hz [1]. Conventional commercial US systems utilize rudimentary beamforming algorithms, such as delay-and-sum (DAS), to reconstruct B-mode images from raw signals, often sacrificing quality for rapidity. The US image reconstruction procedure can be cast as solving an inverse problem. Numerous methods, e.g., adaptive beamforming, have been introduced to enhance US image quality [2, 3]. Concurrently, there is a growing inclination towards harnessing the capabilities of deep learning (DL) for US image reconstruction. These methods, often exhibit superior performance under analogous conditions, which are trained using pairs of pristine images and corresponding measured data [4-6]. However, they tend to exhibit limited generalization capabilities, and they are prone to introduce or exaggerate certain learned image attributes or features [7].

Leveraging data-driven image priors, conceptualized through generative models, has led to substantial progress in tackling challenges in medical image reconstruction, even in the absence of paired datasets. Serving as learned priors derived from intrinsic data distributions, diffusion models have been used within a plug-and-play framework to perform US image reconstruction [8-10]. Nonetheless, the iterative generation process inherent to diffusion models often demands extensive computational resources, a thousand times for sample generation [11, 12]. This results in prolonged inference time, thereby constraining their applicability in real-time scenarios.

In this work, we advocate for the application of the diffusion model to ascertain and probe the inherent distribution pertinent to single PWUS image reconstruction. Specifically, we employ the configurations of diffusion-based generative models (EDM) as delineated in [13], which introduce a novel design paradigm characterized by accelerated sampling. In particular, we enforce a data consistency measure at each sampling phase, contingent upon the single plane wave (PW) measured signals. Furthermore, we dynamically modulate the step size using a sine schedule throughout the iterative process. To the best of our knowledge, this represents the inaugural EDM framework tailored for PWUS image reconstruction. By training on a dataset comprising spatial coherent compounded B-mode images with 75 angles PWs, the EDM is equipped to discern the distribution. We empirically showcase the efficacy of our method on both phantom and *in-vivo* datasets. When provided with single PW data, our method can swiftly sample a high-resolution image from the learned distribution in a small number of steps (e.g., 50 steps). Notably, in terms of quantitative assessments, most of our results even surpass the image compounded with 75 PWs.

## 2. BACKGROUND

### 2.1. Diffusion Generative Models

The forward diffusion process incrementally introduces perturbations to the data by incorporating Gaussian noise. During the training phase, diffusion models are employed to learn the inverse steps through a series of noising iterations. Once trained, the model can be utilized to sequentially generate samples starting from noise.

Let us denote the probability density function of expected data $x_0$ by $p_0(x)$, and $x_t$, obeyed distribution $p_t(x)$, denotes the data at step $t$ in trajectory. The stochastic differential equation (SDE) delineates the diffusion process as outlined in [11]. This equation specifies the solution trajectories originating from $x_t$, utilizing the "probability flow" concept within the framework of an ordinary differential equation (ODE):

---

[†] These authors contributed equally to this work.

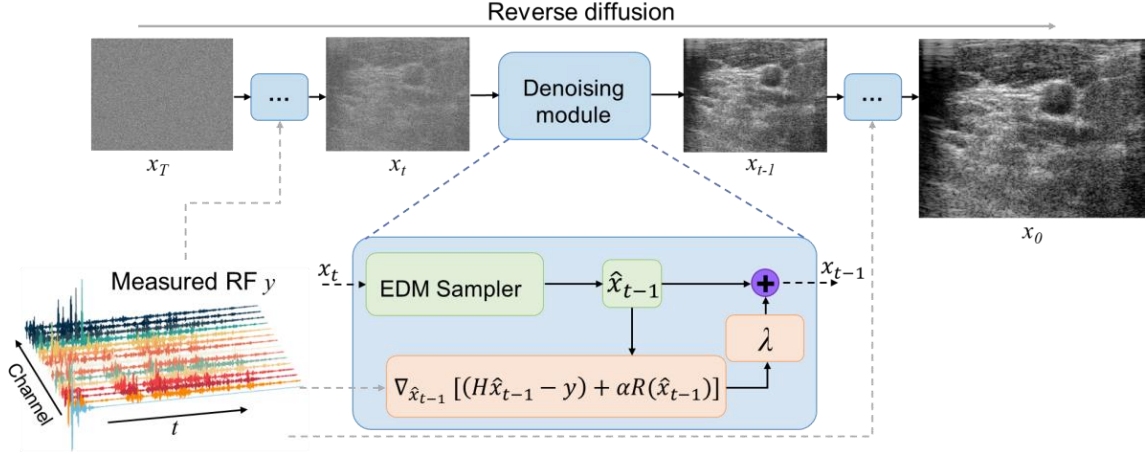

**Fig. 1.** The illustration of EDM-based conditional sampling, which performs an iterative denoising operation to solve US image inverse problems.

$$dx_t = \left[\mu(x_t,t) - \frac{1}{2}\sigma(t)^2 \nabla \log p_t(x_t,t)\right] dt. \quad (1)$$

The score function $\nabla \log p_t(x_t, t)$ points towards higher density of data, which can be leaned by a deep neural network (DNN). Following [13], $\mu(x_t,t) = 0$ and $\sigma(t) = \sqrt{2t}$. For the forward process, we can obtain any $p_t(x) = p_0(x) * N(0, t^2 I)$ $(t \in [0,T], T > 0)$, where $*$ denotes the convolution operation. Given an initial data $\hat{x}_T \sim N(0, T^2 I)$, we solve ODE backwards in time to obtain $\{\hat{x}_t\}_{t \in [0,T]}$ with numerical ODE solver [11]. In general, we view $\hat{x}_0$ as a sample from the data distribution $p_0(x)$. Nonetheless, the sampling procedure necessitates recurrent evaluations of the score function, leading to considerable computational overhead. Karras et al. implemented the higher-order Runge-Kutta method, resulting in a substantial reduction in the number of sampling steps ($< 50$) [13]. This advancement facilitates US image reconstruction through a rapid sampling process, eliminating the need for progressive distillation.

### 2.2. Diffusion Models for US image reconstruction

In PWUS imaging, the received radio-frequency (RF) data $y \in \mathbb{R}^r$ and the desired beamformed data $x_0 \in \mathbb{R}^l$ can be formulated as a linear relationship as:

$$y = H x_0 + z, \quad (2)$$

where $H \in \mathbb{R}^{r \times l}$ is the known measurement matrix, and $z \in \mathbb{R}^r$ denotes additive noise. In this problem, supplementary conditioning information $y$, is linked to $x_0$ through Eq. (2). Our objective is using generative model to learn $p_0(x)$, which indicates a distribution of $x_0$, the image compounded with 75 PWs. The retrieval of $x_0$ is contingent upon $y$, as the objective is to incorporate $y$ through the posterior distribution, represented as $p_0(x_0|y)$. This can be addressed by estimating the problem-specific score as follows:

$$\nabla \log p_t(x_t \mid y) = \nabla \log p_t(x_t) + \nabla \log p_t(y \mid x_t). \quad (3)$$

Noting that the initial term can be approximated using a DNN. This approximation can be achieved by calculating the gradient of the likelihood, represented as $\nabla \log p_t(y|x_t)$. Furthermore, the second term in Eq. (3) can be integrated with the sampling procedure to facilitate image reconstruction.

## 3. METHODS

### 3.1. Fast Sampling in EDM

Key areas of interest in the diffusion model encompass enhancing the output quality and/or reducing the computational expense associated with sampling. Previous studies determined the time steps during the training phase, which consequently fixed both $\sigma(t)$ and the scaling $s(t)$. Karras et al. highlighted that by solely considering the parameter range during training, the sampling procedure can be rendered more adaptable [13]. Specifically, it becomes feasible to generate any points along the solution trajectory during training. In sampling process, one can opt for varying time steps to achieve optimal quality. This quality should diminish in a consistent manner with a decreasing $\sigma$, as illustrated by:

$$\sigma_{t>0} = \left( \sigma_{\min}^{\frac{1}{\rho}} + \frac{t-1}{T-1} (\sigma_{\max}^{\frac{1}{\rho}} - \sigma_{\min}^{\frac{1}{\rho}}) \right)^{\rho}, \sigma_0 = 0. \quad (4)$$

The exponential scale $\rho$ can adjust the steps near $\sigma_{\min}$ and $\sigma_{\max}$, and we empirically use $\rho = 7$ in this work. Then, we use Heun's second-order method to sample the image from desired distribution.

### 3.2. Conditional Sampling for Single PW Reconstruction

Our objective, with the support of EDM, is to derive a high-quality US image using a single PW. Building upon the aforementioned method, we have accomplished random sampling in a reduced number of steps. In comparison to previous sampling techniques, EDM has the capability to curtail the step count to fewer than 50.

The subsequent phase involves approximating the likelihood, $p_t(y/x_t)$, by integrating the data consistency step.

Fig. 1 provides a visual representation of the sampling process utilizing EDM. The EDM sampler signifies the application of Heun's second-order sampling step, initiated with random noise. Following each sampling iteration, it becomes imperative to incorporate the provided RF signals, serving as data-consistent guidance as depicted in Fig. 1. This is determined by the gradient of the discrepancy between the sampled outcome and the observed signals. During this process, certain priors pertinent to the US image can be introduced as a regularization term, as detailed below:

$$x_{t-1} = \hat{x}_{t-1} - \lambda \nabla_{\hat{x}_{t-1}} [\|H\hat{x}_{t-1} - y\|_2 + \alpha R(\hat{x}_{t-1})], \quad (5)$$

where $R$ denotes the regularization term, $\alpha$ is a scale of regularization, while $\lambda$ is used to balance the contribution of data consistency. It is imperative to highlight that the $\lambda$ value should be minimal during the start of sampling, given that the sampling commences from noise. If not moderated, this could amplify the truncation error associated with the ODE solver, subsequently leading to convergence issues in the sampling process. To circumvent this, we employ a sine-varying scheduler to diminish the weight at both the beginning and end of the sampling. Such an approach ensures a stable and reliable reconstruction.

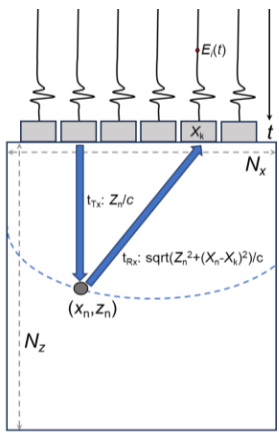

**Fig. 2.** The established procedure of the measurement matrix $H$.

### 3.3. Implementation Details

We establish the measurement matrix used in the sampling process in Fig. 2 [14]. A constant speed $c$ of US (1540 m/s) and nonexistence of multiple scattering on an $N_z \times N_x$ (1000×256) grid are assumed. For each pixel, we compute the time of flight for the US wave from each element of the US transducer, spanning from its transmission moment to its reception by another element. This is then translated into a position on the RF signal. The measurement matrix $H$ ought to be pre-determined for each US transducer ($N_e$=128 elements, 25 MHz sampling rate, $N_s$=1300 samples per channel) and imaging sequence as a sparse matrix with a size of $(N_z \times N_x) \times (N_e \times N_s)$.

In our work, beamformed data with 75 PWs were used for training the denoising model to learn data distribution. The EDM framework was implemented in Pytorch [15], which was trained on a workstation equipped with 128 GB RAM and four NVIDIA TITAN V GPUs. In the sampling process, the maximum time step was 50 in this work.

## 4. EXPERIMENTAL RESULTS

### 4.1. Phantom and *in-vivo* Data

Our experimental demonstrations were conducted on both tissue-mimicking phantom and *in-vivo* datasets. The data were obtained with an L10-5 linear-array transducer connected to a Vantage system (Verasonics Inc., Kirkland, WA, USA). This system captured images comprising 75 steered PWs, uniformly distributed between angles of -16° and 16°. For the *in-vivo* dataset, 300 frames of channel RF data were acquired from the right carotid artery of two healthy volunteers, in both longitudinal and cross-sectional views. Additionally, a tissue-mimicking phantom (Model 403 LE, Gammex, Middleton, WI, USA) was scanned to obtain 100 frames of channel RF data. A test dataset was subsequently assembled using the acquired 100 frames of RF data for reconstruction.

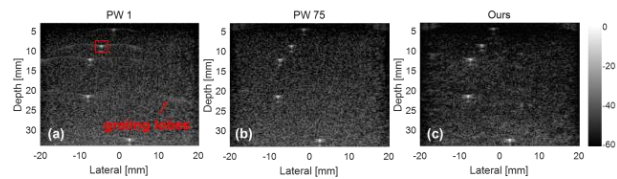

**Fig. 3.** B-mode images of the point-reflector phantom reconstructed using (a) DAS with a single PW, (b) DAS with 75 PWs, and (c) the proposed approach with a single PW, respectively.

### 4.2. Phantom Results

In our experiments, we compared the results of three methods, i.e., DAS with a single PW, DAS with 75 PWs, and the proposed method with a single PW. The B-mode images of the point-reflector phantom are shown in Fig. 3. As illustrated by Figs. 3(a) and (b), spatial coherent compounding is adept at mitigating the side lobes within the image. The proposed method [Fig. 3(c)] demonstrates superior background noise suppression when contrasted with DAS using a single PW [Fig. 3(a)].

**Table 1.** Mean axial and lateral FWHMs (mm) of the point reflector in the red region of the point-reflector phantom for different methods

| FWHM | DAS (1 PW) | DAS (75 PWs) | Ours (1 PW) |
|---|---|---|---|
| Axial | 0.3470 | 0.3384 | 0.3581 |
| Lateral | 0.4227 | 0.4214 | 0.4255 |

The mean full widths at half maximum (FWHMs) of the point reflector in the red region in both axial and lateral directions for the three methods are listed in Table 1. The results reveal that the FWHMs for the three methods are similar. The results of our method, akin to those obtained by DAS with 75 PWs, excel in sidelobe suppression without compromising the resolution.

### 4.3. *In-vivo* Results

Fig. 4 shows the B-mode images reconstructed using both DAS and the proposed method. A discernible observation is that the proposed method yields images with enhanced contrast while adeptly maintaining the integrity of tissue structures, such as the vessel intima, especially when compared with the DAS result with a single PW. In the context of DAS with a single PW, the pronounced noise significantly diminishes the vessel contrast, making it particularly challenging to discern the carotid artery in the cross-sectional view. However, as the number of compounded PWs escalates to 75, there is a noticeable suppression of this noise. Consequently, the carotid lumen becomes distinguishable in both longitudinal and cross-sectional views. Our method shows similar results to DAS with 75 PWs while using only 1 PW.

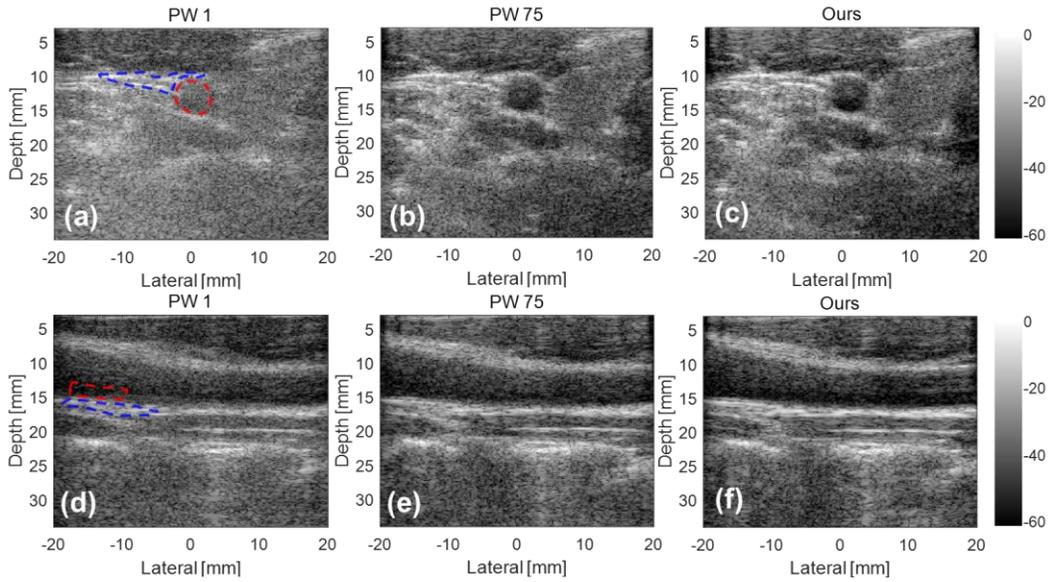

**Fig. 4.** B-mode images of the common carotid artery of a volunteer reconstructed from DAS with [(a) and (d)] 1 PW and [(b) and (e)] 75 PWs and [(c) and (f)] the proposed method with 1 PW, respectively. The [(a)-(c)] first and [(d)-(f)] second rows are from the cross-sectional and longitudinal views of the common carotid artery, respectively.

**Table 2.** Quantitative results (CNR in dB and gCNR) of the common carotid artery *in-vivo* for different methods

|  |  | DAS (1 PW) | DAS (75 PWs) | Ours (1 PW) |
|---|---|---|---|---|
| Cross-sectional | CNR | 1.8890 | 2.2844 | **2.7480** |
|  | gCNR | 0.7656 | 0.8921 | **0.9151** |
| Longitudinal | CNR | 5.1481 | **6.1987** | 6.0090 |
|  | gCNR | 0.9683 | 0.9897 | **0.9935** |

We manually selected ROIs in the vessel wall (blue dashed lines) and lumen (red dashed lines) of the carotid artery in different views to compute the contrast-to-noise ratio (CNR) and the generalized contrast-to-noise ratio (gCNR) [16], as presented in Table 2. The proposed method with a single PW obtains higher CNR and gCNR than DAS with 75 PWs.

## 5. CONCLUSION

In this paper, we introduce an EDM-based reconstruction method to enhance the image quality of PWUS. By assimilating a prior based on the B-mode image compounded with 75 PWs, our proposed method with a single PW obtains an image quality similar to that of DAS with 75 PWs. Furthermore, in the inference stage, the proposed approach can produce high-fidelity images in only 50 iterations. Our experimental findings indicate that, in most *in-vivo* cases, the proposed method with a single PW even outperforms DAS with 75 PWs. In the future, we will validate the efficacy of our approach with a more extensive dataset and strive to further curtail the iteration count.

## 6. COMPLIANCE WITH ETHICAL STANDARDS

Approval of all ethical and experimental procedures and protocols was granted by the Medical Ethics Committee of Tsinghua University.

## 7. ACKNOWLEDGMENTS

The authors would like to extend their thanks to Shangqing Tong (at ShanghaiTech University) for his invaluable support and contribution towards the methodology and discussion of this paper. This work was supported in part by the National Natural Science Foundation of China (62171442 and 62027901) and Beijing Natural Science Foundation (M22018).